\documentclass[a4paper]{article}

\usepackage{INTERSPEECH2021}
\usepackage{amsmath,graphicx}
\usepackage{xcolor}
\usepackage{color, soul, multirow}
\usepackage{amssymb}
\usepackage{subfig}
\usepackage{bm}
\usepackage{verbatim}
\usepackage{cite}
\usepackage{amssymb}
\usepackage{pifont}
\usepackage{tabularx}
\usepackage{lipsum}
\usepackage{float}
\usepackage{booktabs}
\usepackage{float}
\usepackage{tikz}
\usepackage{tikz-qtree}
\usepackage{hyperref}

\title{Semantic Distance: A New Metric for ASR Performance Analysis\\
Towards Spoken Language Understanding}
\name{
Suyoun Kim, Abhinav Arora, Duc Le, Ching-Feng Yeh, \\ Christian Fuegen, Ozlem Kalinli, Michael L. Seltzer 
}
\address{
  Facebook AI}
\email{suyounkim@fb.com}

\begin{document}

\maketitle
\begin{abstract}
    Word Error Rate (WER) has been the predominant metric used to evaluate the performance of automatic speech recognition (ASR) systems. However, WER is sometimes not a good indicator for downstream Natural Language Understanding (NLU) tasks, such as intent recognition, slot filling, and semantic parsing in task-oriented dialog systems. This is because WER takes into consideration only literal correctness instead of semantic correctness, the latter of which is typically more important for these downstream tasks. In this study, we propose a novel Semantic Distance (SemDist) measure as an alternative evaluation metric for ASR systems to address this issue. We define SemDist as the distance between a reference and hypothesis pair in a sentence-level embedding space. To represent the reference and hypothesis as a sentence embedding, we exploit RoBERTa, a state-of-the-art pre-trained deep contextualized language model based on the transformer architecture. We demonstrate the effectiveness of our proposed metric on various downstream tasks, including intent recognition, semantic parsing, and named entity recognition.
  
\end{abstract}

\section{Introduction}
    
    While the adoption of Word Error Rate (WER) as the de facto evaluation metric has served to advance automatic speech recognition (ASR) research over the decades, there has been an increasing interest in the speech recognition community to consider a more suitable evaluation measure for downstream Natural Language Understanding (NLU) applications, such as intent recognition, slot filling and semantic parsing for task-oriented dialog. This is primarily because WER has been shown to have limitations in measuring semantic correctness, as it is derived from the word-level edit distance between the true transcription and the ASR hypothesis, where every error (substitution, insertion, or deletion)
    %in the same category (substitution, insertion, or deletion) 
    is weighted equally. For example, if the reference is ``\emph{This is a cat}'' and two ASR systems generate different hypotheses: ``\emph{This is the cat}'' and ``\emph{This is a cap}'', then the former system would be preferred by a downstream NLU system. However, WER by itself cannot identify which system is better as the error rates are identical (one substitution error). Past research has highlighted such limitations of WER and demonstrated that improvements in NLU can be obtained while observing a worse WER \cite{wang2003word, garofolo2000trec, grangier2003information}.  
    
    Motivated by the limitations of WER, alternative measures have been proposed. \cite{morris2004and} presented word information preserved (WIP) based on mutual information between the reference and the hypotheses. \cite{garofolo19991998} proposed a new measure that includes named Entity Error Rate (EER), and the stop-word-filtered WER, for taking word importance weight into account. \cite{makhoul1999performance, hunt1990figures, mccowan2004use} attempted to adopt information retrieval to measure the performance. While these metrics addressed some of WER's limitations, all of them are still based on the literal-level word correctness and do not allow for direct analysis of performance at the semantic level of the sentence.
    
    Recently, substantial work has shown that pre-trained neural language models, trained on billions of words, can learn universal language representations of text in the form of low-dimensional continuous feature vector (i.e., embedding) in the semantic space. These embeddings can then be plugged into a variety of downstream tasks, such as textual similarity, question answering, paraphrasing, sentiment analysis, etc., to drastically improve their performance \cite{Peters:2018,devlin2018bert,liu2019roberta}. In 2017, \cite{Peters:2018} introduced 
    ELMo and demonstrated that contextualized word representations from this model outperformed earlier word embeddings such as Word2Vec \cite{mikolov2013distributed} and GloVe \cite{pennington2014glove} by capturing linguistic context in addition to word-level syntax and semantics. The GPT model \cite{Radford2018ImprovingLU} proposed  a new architecture using transformers \cite{vaswani2017attention} and was used for text generation tasks. BERT, which is based on bidirectional transformer, set a new state-of-the-art performance on 11 NLU tasks \cite{devlin2018bert}. Later, RoBERTa subsequently showed that BERT can be further improved by robustly optimized the pre-training process \cite{liu2019roberta}. More importantly, BERT and RoBERTa have demonstrated their ability to derive semantically meaningful sentence embeddings that can be compared using cosine similarity \cite{reimers2019sentence}. To the best of our knowledge, there have been no studies thus far that leverage these models to evaluate the performance of ASR systems.

    In this work, we propose a novel Semantic Distance (SemDist) measure as an alternative performance metric for ASR systems to capture semantic correctness. We define Semantic Distance as the distance between the reference and an ASR hypothesis in the sentence embedding semantic space. To represent the reference and hypothesis as a sentence embedding, we exploit RoBERTa \cite{liu2019roberta}, a state-of-the-art pre-trained deep contextualized language model based on the transformer architecture. We evaluate SemDist on several downstream tasks, including intent recognition, semantic parsing, and named entity recognition. We demonstrate that our proposed metric has better correlation with NLU performance than WER and can potentially be used as part of the model selection process.

    \begin{figure}[t]
      \centering
      \includegraphics[width=1\columnwidth]{ 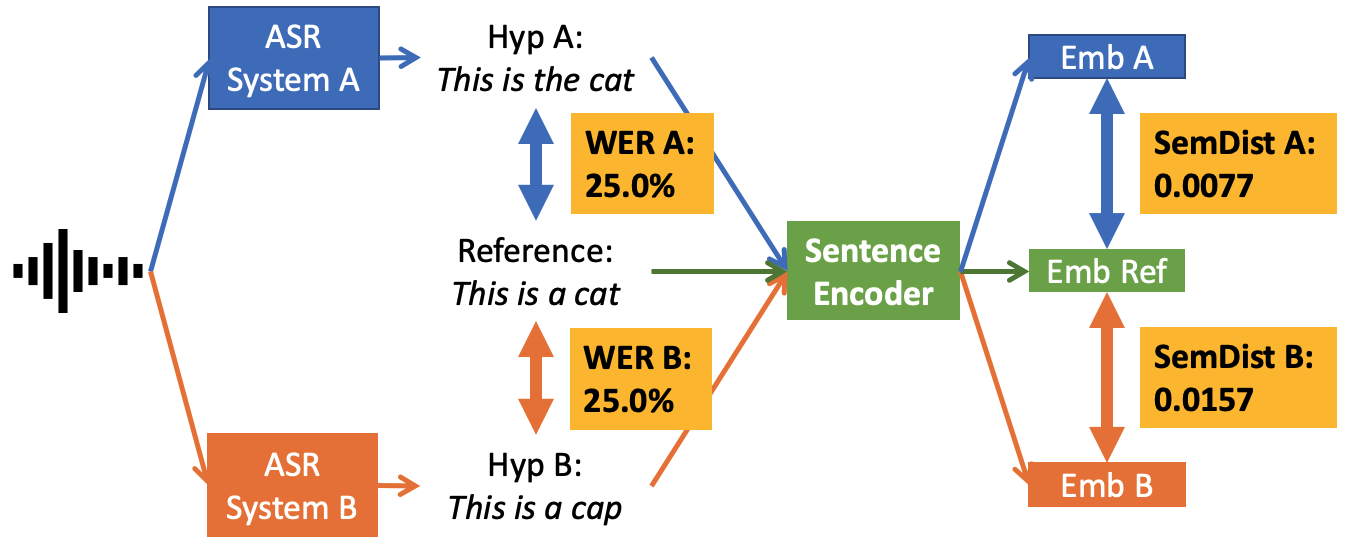}
      \caption{Two metrics for evaluating ASR systems: the WER and our proposed SemDist.}
      \label{fig:sem_dist_score}
    \end{figure}
    
\section{Semantic Distance}
\label{sec:method}
    In this section, we describe our proposed SemDist as an alternative ASR performance metric. Figure~\ref{fig:sem_dist_score} illustrates the overall procedure to obtain the SemDist from the ASR systems A and B in addition to the WERs. Our proposed SemDist is calculated in two steps. First, we exploit pre-trained sentence embedding models to map the utterances into a sentence embedding space (in Section \ref{sec:roberta}). Second, we measure semantic distance by using the cosine similarity function (in Section \ref{sec:cos}).
    
\subsection{Sentence Embeddings}
\label{sec:roberta}
    To represent the reference and hypothesis in the sentence-level semantic embedding space, we use RoBERTa \cite{liu2019roberta}, a state-of-the-art pre-trained masked language model. It uses the same architecture as BERT \cite{devlin2018bert}, with the robustly optimized training method, (i.e. longer training, with bigger batches,  dynamically changing the masking pattern, etc), described in \cite{liu2019roberta}. It has produced state-of-the-art results on a wide variety of challenging NLP benchmarks, such as GLUE, SQUAD and RACE. RoBERTa/BERT employ bidirectional training of transformers\cite{vaswani2017attention}, allowing the models to learn a deeper sense of language context. Further, with the language masking strategy used for training, the model learns to predict intentionally masked sections of text. Most importantly, open source RoBERTa and BERT models, pre-trained on billions of words, are readily available and allow fast fine-tuning with small modifications (i.e. additional output layer) on any specific final task. This form of transfer learning, where pre-trained models are used as starting points for task-specific models, has shown significant breakthroughs in semantic textual similarity tasks\cite{xlnet}.
    
    In our method, we pass the reference and hypothesis through the pre-trained RoBERTa and perform a pooling operation by computing the mean of all output vectors. The pre-trained model architecture that we used in this work is $\text{BERT}_{\text{BASE}}$ \cite{devlin2018bert}, which has 12 transformer layers with 768 hidden size and 12 self-attention heads, for a total of 110M parameters. Thus, a single single 768-dimensional sentence embedding vector is computed for each reference and hypothesis.
    
    % In addition to RoBERTa, we also experiment with out of the box LASER\cite{laser}\footnote{\href{https://github.com/facebookresearch/LASER}{github.com/facebookresearch/LASER}} embeddings of 1024 dimension. These are generated by a multilingual BiLSTM encoder, which is trained using the framework of neural machine translation. Since these are multilingual embeddings, they can also be readily used for SemDist calculations in other languages.
    
\subsection{Cosine Distance Scoring}
\label{sec:cos}
    In the sentence embedding space, a simple cosine distance has been applied successfully to compare two utterances for semantic textual similarity decision \cite{reimers2019sentence}. Given two sentence embeddings: an embedding of the reference transcription, $e_{\text{ref}}$, and an embedding of the hypothesis generated from an ASR system, $e_{\text{hyp}}$, SemDist is calculated as follows:
    \begin{equation}
        \texttt{SemDist}( e_\text{ref}, e_\text{hyp} ) = 1 - \frac{(e_\text{ref})^T\cdot e_\text{hyp}}{|| e_\text{ref} || \cdot || e_\text{hyp} ||  }
    \end{equation}
    Note that the cosine distance only considers the angle between the two sentence embeddings and not their magnitudes. SemDist is bounded between 0 and 1, where lower scores indicate higher semantic similarity and vice versa. %It is believed that the length of sentence affects the sentence embedding magnitudes so removing magnitude in scoring greatly improves the robustness of the semantic similarity measure.

    \begin{table}[th]
      \caption{Example comparison of our proposed SemDist and other conventional metrics, WER, NER, and POS tagging accuracy of two hypotheses from different ASR systems. The reference transcription is ``This is a cat."}
      \label{tab:cat}
      \resizebox{1\columnwidth}{!}{
          \centering\small\begin{tabular}{rrrrrr}
          \toprule
            \textbf{ASR} & \textbf{Hypo.}  & \textbf{WER} & \textbf{NER} & \textbf{POS} & \textbf{SemDist} \\
          \midrule
            A & \textit{This is the cat}  & 25.0\% & None & 100\% & 0.0077 \\
            B & \textit{This is a cap} & 25.0\% & None & 100\% & 0.0157 \\
          \bottomrule
          \end{tabular}
      }
    \end{table}
    
    Table~\ref{tab:cat} demonstrates the difference between our proposed SemDist and other conventional metrics, WER, Named Entity Recognition F1-score (NER), POS tagging accuracy (POS) on two different ASR hypotheses, given the reference transcription ``\emph{This is a cat}''. Naturally, the downstream tasks prefer Model A to Model B because Model A is more semantically correct. As seen in this example, WER and other metrics cannot separate these two models since it only measures literal word-level correctness. However, SemDist can indicate that Model A (0.0077) performed better than Model B (0.0157).  

\section{Experimental Setup}
\label{sec:exp}
    \subsection{Overall Experiment Pipeline}
    \label{sec:hypo}
    In order to demonstrate the effectiveness of our proposed SemDist metric on various realistic large-scale downstream tasks, we use strong baseline ASR and NLU systems, and evaluate on our large-scale in-house ASR/NLU dataset. We describe these baseline systems and evaluation dataset in more detail in Section~\ref{sec:asr_model}, Section~\ref{sec:nlu_model}, and Section~\ref{sec:eval_data}, respectively. 
    
    In order to address our main research question ``\textit{Can SemDist identify which ASR system is better even when WER is same?}'', we derive three more ASR outputs, hypotheses sets, in addition to our ASR baseline output. We then evaluate and compare the performance of these hypotheses set in NLU task by using NLU metrics (in Section~\ref{sec:nlu_metric}). 
    
    We first obtain the hypotheses (Set A) from our strong ASR baseline on the evaluation dataset. 
    We then generate Set B that has the exactly same WER, but has worse (higher) SemDist. To do so, based on each hypothesis' number of substitution/insertion/deletion errors in Set A, we substitute or insert the true/reference word with a random word or delete the random position of the reference word. 
    On the contrary, we generate Set C that has the exactly same WER, but has better (lower) SemDist. To do so, we change the order of two random reference word or add the articles (i.e. ``\textit{a}'') to minimize the damage of the meaning of the reference sentence. 
    Finally, we also generate Set D that has better (lower) SemDist without limiting to have same WER, but also without artificial way. To do so, we use a neural language model and perform the shallow fusion \cite{Kim2021lmfusion} with internal LM subtraction \cite{Meng2021ILME} on top of our strong ASR baseline. This results in better WER as well, but we can get more realistic insight on SemDist. 
    
    \subsection{ASR Task}
    \label{sec:asr_model}
        We first build a strong baseline ASR system by employing a large-scale in-house ASR training dataset consisting of two parts. The first part comprises of 1.7M hours of English video data publicly shared by Facebook users; all videos are completely de-identified before transcription, and both transcribers and researchers do not have access to any user-identifiable information (UII). The second part contains approximately 50K hours of manually transcribed de-identified data with no UII in the voice assistant domains. 
        
        Our ASR model is an end-to-end sequence transducer, a.k.a. RNN-T \cite{Graves12transduction} with approximately 83M total parameters. The acoustic encoder is a 20-layer streamable low-latency Emformer model \cite{Shi2021emformer} with a stride of 6, 60ms lookahead, 300ms segment size, 512-dim input, 2048-dim hidden size, eight self-attention heads, and 1024-dim fully-connected (FC) projection. The text predictor consists of three Long Short Term Memory (LSTM) layers with 512-dim hidden size, followed by 1024-dim FC projection. The joiner network contains one Rectified Linear Unit (ReLU) and one FC layer. The target units are 4095 unigram WordPieces \cite{Kudo2018SubWord} trained with SentencePiece \cite{kudo-richardson-2018-sentencepiece}. The model is first trained for 4 epochs using sub-word regularization ($l=5$, $\alpha=0.25$) \cite{Kudo2018SubWord}, SpecAugment LD policy \cite{park2019specaugment}, and AR-RNNT loss \cite{Mahadeokar2021AR-RNNT} (left buffer 0, right buffer 15), where the alignment is provided by a chenone hybrid acoustic model (AM) \cite{Le2019Kulfi}. Finally, we fine-tune the model for 1 epoch with trie-based deep biasing \cite{Le2021deepshallow}.
            
    \subsection{Evaluation Dataset}  
    \label{sec:eval_data}
        Our in-house annotated evaluation sets for ASR task have two main domains: open-domain dictation and assistant-domain voice commands. 
        The open-domain dictation set includes 22.9K de-identified utterances (305K words) collected from crowd-sourced workers on mobile devices. It contains a mix of short-form (9.3 words/utterance) and long-form (19.0 words/utterance) dictation data under diverse recording environments. 
        The assistant-domain voice commands set includes 15K manually transcribed de-identified utterances (46K words) collected from voice activity of volunteer participants. The participants consist of households that have consented to have their voice activity reviewed and analyzed. We use these datasets to analyze the relationship between WER and SemDist, and basic NLP metrics (in Section~\ref{sec:res_asr} and Section~\ref{sec:res_ner}). 
        
        For NLU, annotated evaluation sets have only assistant-domains, and there are 10k utterances that overlap with the ASR assistant-domain evaluation set. We use this dataset (NLU $\cap$ ASR) for the tasks of intent recognition and semantic parsing (in Section~\ref{sec:res_nlu}). 
        
        \begin{table}[th]
            \caption{Description of ASR/NLU task evaluation dataset}
            \label{tab:data}
            \centering
            \resizebox{0.9\columnwidth}{!}{
            \centering
                \begin{tabular}{rrrrrr}
                \toprule
                \textbf{Task}& \textbf{Domain}  & \textbf{\# utter} & \textbf{\# word} & \textbf{Avg. Len.} \\
                \midrule
                ASR & Open      & 23k  & 305k  & 13.3         \\
                ASR & Assistant & 15k  & 46k   & 3.0         \\
                NLU $\cap$ ASR & Assistant & 10k  & 25k   & 2.4        \\
                \bottomrule
            \end{tabular}
            }
        \end{table}

    \subsection{NLU Task: Intent Recognition, Semantic Parsing}
    \label{sec:nlu_model}
\begin{figure}[h]
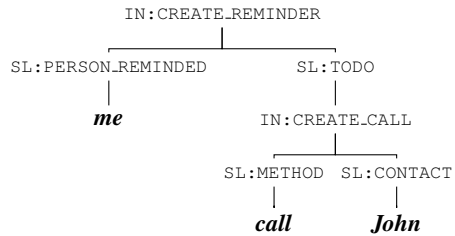

        \centering\small
        \tikzset{level distance=20pt,sibling distance=0pt}
        \tikzset{edge from parent/.style={draw, edge from parent path={(\tikzparentnode.south) -- +(0,-8pt) -| (\tikzchildnode)}}}
        \Tree [.\texttt{\scriptsize IN:CREATE\_REMINDER} [.\texttt{\scriptsize SL:PERSON\_REMINDED} \emph{{\textbf{me}}} ] [.\texttt{\scriptsize SL:TODO} [.\texttt{\scriptsize IN:CREATE\_CALL} [.\texttt{\scriptsize SL:METHOD} \emph{{\textbf{call}}} ] [.\texttt{\scriptsize SL:CONTACT} \emph{{\textbf{John}}} ] ] ] ]
        \caption{Decoupled semantic representations for the single utterance ``Please remind me to call John''.}\label{fig:decoupled-example}
        \vspace{-0.3cm}
        \end{figure}

        For NLU downstream task, we evaluate four sets of assistant-domain hypotheses (A/B/C/D) with different SemDist on Intent Recognition and Semantic Parsing\cite{GuptaSMKL18} tasks. Intent recognition is a text classification task where we predict the top-level intent of the utterance from a set of 351 intent types. For the semantic parsing task, we use the recently introduced decoupled semantic representation form\cite{AghajanyanMSDHL20}. The decoupled representation allows for compositional semantic structures, where a slot can further contain nested intents and slots within itself, providing high expressiveness for task-oriented dialog systems. Figure~\ref{fig:decoupled-example} shows an example of the decoupled representation for the the utterance ``\textit{Please remind me to call John}'', which has \texttt{IN:CREATE\_REMINDER} as the the top-level intent. %The decoupled representation can be serialized 
        
        %For NLU evaluation, we build a test set in the same way and combine it with the assistant-domain voice commands evaluation set, discussed in ~\ref{lab:asr_dataset}, to obtain an evaluation set consisting of 10,405 \textit{(audio, transcript, NLU annotation)} triplets. 
        
        We build a strong baseline NLU system by using an internal dataset, which consists of about 475k annotated utterances, across 38 domains and 351 intents. The utterances in this dataset were generated via crowd-sourcing and were manually labelled by annotators using the process described in \cite{GuptaSMKL18}. 
        
        Our NLU model is a sequence-to-sequence architecture, described in \cite{AghajanyanMSDHL20}, where the source sequence is the utterance and the target sequence is the serialized decoupled representation. At every decoding step, the model can either generate a token from the intent-slot ontology, or copy a token from the source sequence via a pointer-generator mechanism. The model uses two distinct stacked bidirectional LSTMs\cite{hochreiter1997long} as the encoder and stacked unidirectional LSTMs as the decoder. Both consist of two layers of size 512, with randomly initialized embeddings of size 300. We also incorporate contextualized word vectors, by augmenting the input with ELMo embeddings\cite{Peters:2018}.
        
    \subsection{NLP Task: Named Entity Recognition}
    \label{sec:ner_task}
        In addition to the above NLU tasks, we also evaluate four sets of open-domain hypotheses (A/B/C/D) with different SemDist on a Named Entity Recognition (NER) task. Recognition of named entities such as names of people, organizations, locations, etc, is often used to understand the meaning of text. Thus, we investigate how SemDist relates to the performance on a NER task. Since our dataset does not have annotated entities, we use an open-source software library Spacy \cite{spacy} to generate the entities of the reference transcriptions and use them as pseudo labels. We then generate the entities for each hypotheses set and measure the F1-score. 
        
    \subsection{Metrics for Downstream Tasks}
    \label{sec:nlu_metric}
        To evaluate the NLP/NLU performance in the downstream task, we used four metrics:
        \begin{enumerate}
            % \item \textbf{Domain accuracy}: Percentage of utterances where the NLU domain of the prediction is same as the ground truth domain. 
            \item \textbf{Intent accuracy (IntentAcc)}: Percentage of utterances where the top-level intent in the decoupled form in the prediction matches the ground truth. 
            \item \textbf{Exact match accuracy (EM)}: Similar to \cite{GuptaSMKL18}, we define exact match accuracy as the percentage of utterances where the complete decoupled form is correct. This is the strictest metric, which is 1 only when all the intents and the slots in the utterance are predicted correctly.
            \item \textbf{Exact match tree accuracy (EM Tree)}: One drawback of the EM metric is that it will always be 0 when ASR makes a mistake in recognizing slot tokens. Therefore, to study the effectiveness of NLU in the light of such mistakes, we also evaluate the exact match accuracy of the decoupled form after dropping the slot text, which allow us to identify the percentage of utterances where NLU was able to identify the correct semantic frame, regardless of ASR errors in recognizing slot tokens.
            \item \textbf{NER-F1}: F1-score of the predicted named entities of the ASR hypotheses
        \end{enumerate}
       
\section{Results and Discussions}
    \subsection{Correlation between WER and Semantic Distance}
    \label{sec:res_asr}
\begin{figure}[H]
          \centering
          \includegraphics[width=0.9\columnwidth]{ 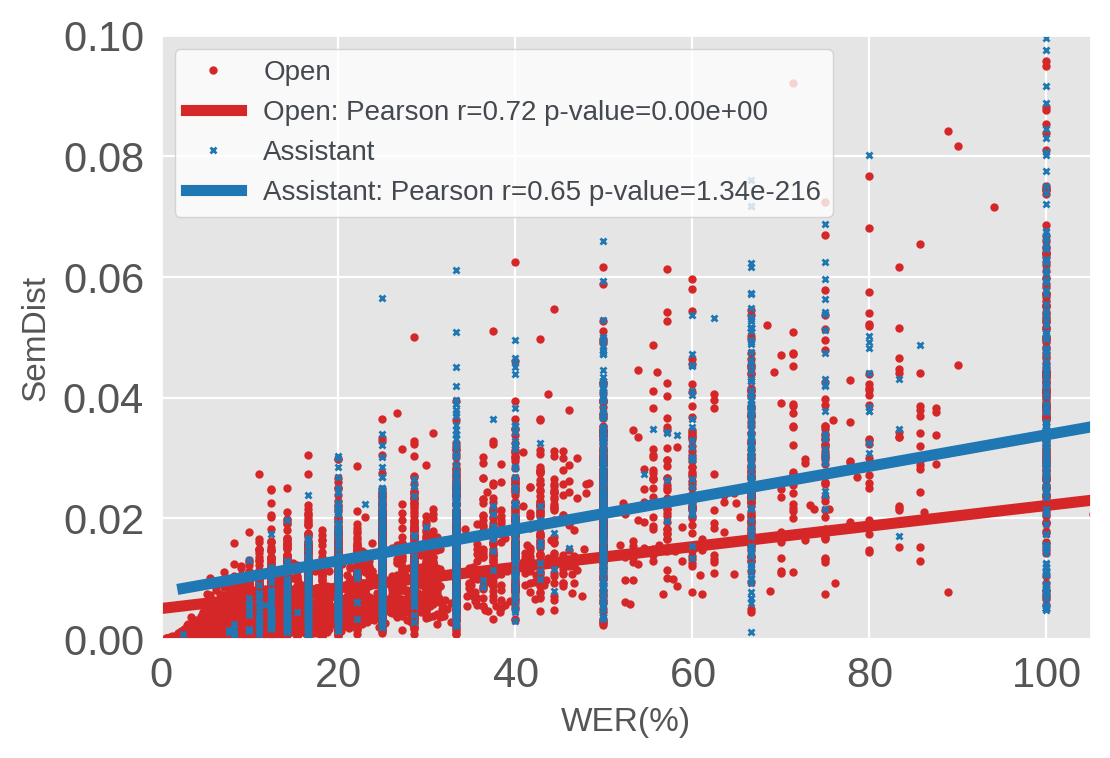}
         \caption{Correlation between SemDist and WER. The red `o' mark represents 7,871 number of open-domain test utterances and the blue `x' mark represents 1,829 number of assistant-domain test utterances that show 0 $<$ WER $<=$ 100\%}
          \label{fig:sem_wer_all}
        \end{figure}

        We first analyze the correlation between our proposed SemDist and WER. As seen in Figure~\ref{fig:sem_wer_all}, we observe that SemDist and WER are highly positively correlated in both open and assistant domains. The Pearson correlation coefficients are 0.72 and 0.65 in the open and assistant domain respectively. In addition, we observe that as WER gets higher it shows more widely spread SemDist at the same WER. This suggests that ``not all errors are equal," and by focusing exclusively on WER we may miss more nuanced differences between the hypotheses.
        
    \subsection{Open-domain: NER and Semantic Distance}
    \label{sec:res_ner}
        \begin{table}[th]
          \caption{Results of WER, SemDist, and NER F1 score}
          \label{tab:ner_results}\centering
          \resizebox{0.8\columnwidth}{!}{
              \begin{tabular}{rrrr}
              \toprule
                \textbf{} & \textbf{WER}  & \textbf{SemDist} & \textbf{NER-F1} \\
              \midrule
                Set A (BS)       & 7.44 & 0.0033& 0.747\\
                Set B (WorseSem) & 7.44 & 0.0044& 0.590\\
                Set C (BetterSem)& 7.44 & 0.0028& 0.846\\
                Set D (BS$+$LM)  & 7.03 & 0.0031& 0.758\\
              \bottomrule
              \end{tabular}
          }
        \end{table}
        \vspace{-0.1cm}
        We next investigate the relationship between the SemDist and the Named Entity Recognition (NER) on the open-domain test set (described in \ref{sec:ner_task}), which has 23k utterances. Table~\ref{tab:ner_results} shows the WER, SemDist, and NER F1-score of four different sets of ASR hypotheses: A(BS)/B(WorseSem)/C(BetterSem)/D(BS$+$LM) (described in \ref{sec:hypo}) on the evaluation set. We observed that as SemDist reduces entity F1-score increases, even with the same WER (Set A vs. Set C). The result indicates that our proposed SemDist measure also aligns with the simple NLP task of NER. Note that the Entity Error Rate(EER) is often used as an additional metric of ASR performance; however, it still has limitations in measuring semantic correctness as seen in Table \ref{tab:cat}.
    
    \subsection{Assistant-domain: NLU tasks and Semantic Distance}
    \label{sec:res_nlu}
        We also analyzed the relationship between SemDist and the NLU task (described in \ref{sec:nlu_model}) on the assistant-domain test set, which has 10k utterances. Similar to the NER experiments (in \ref{sec:res_ner}), we compared NLU metrics of the four different sets of ASR hypotheses: A(BS)/B(WorseSem)/C(BetterSem)/D(BS$+$LM) (described in \ref{sec:hypo}) by using our NLU system, described in \ref{sec:nlu_model}. Table~\ref{tab:nlu_results} shows the WER, SemDist, and NLU metrics: Intent accuracy, EM, and EM Tree (described in \ref{sec:nlu_metric}). We observed that as SemDist reduces, Intent accuracy, EM, and EM Tree increase, even with the same WER (Set A vs. Set C). The results indicate that our proposed SemDist can be a better indicator than WER for various downstream NLU tasks as well.
        
        \begin{table}[th]
          \caption{Results of WER, SemDist, and NLU Metrics}
          \label{tab:nlu_results}
          \resizebox{1\columnwidth}{!}{
              \centering\begin{tabular}{rrrrrr}
              \toprule
                \textbf{} & \textbf{WER}  & \textbf{SemDist} & \textbf{IntentAcc} & \textbf{EM} & \textbf{EM Tree}\\
              \midrule
                Set A (BS)       & 6.16& 0.0024 & 94.63 & 90.81 & 91.34\\
                Set B (WorseSem) & 6.16& 0.0030 & 94.28 & 90.27 & 90.73\\
                Set C (BetterSem)& 6.16& 0.0017 & 96.22 & 92.98 & 93.08\\
                Set D (BS$+$LM)  & 6.01& 0.0023 & 94.84 & 91.14 & 91.58\\
              \bottomrule
              \end{tabular}
          }
        \end{table}
        
    \subsection{Examples}
        Table~\ref{tab:example} shows example hypotheses with their SemDist on the open-domain set. We selected the examples that have same WER on both Set A and D. In the first example, although both hypotheses A and D are incorrect and has same WER, SemDist indicates that D is more semantically close to REF. Since (\textit{`she' vs. `he'}) are at least the same Part-Of-Speech(POS) tag - subject, we expect that our SemDist may be beneficial to the downstream tasks, such as sentence parsing, POS tagging. In the second example, even though A is more similar in pronunciation (\textit{`hitting that' vs. `had not'}), we observed that SemDist is higher in A because it is contradict the REF. We also observed that our SemDist takes semantically more important word (\textit{`aw/or' vs. `aw/oh'}) into account on measuring as seen in the third example. 
        
        \begin{table}[th]
          \caption{Examples of hypothesis with SemDist.}
          \label{tab:example}
            \centering\small
          \resizebox{0.8\columnwidth}{!}{
            \begin{tabular}{rl}
                \toprule
                \textbf{SemDist} & \textbf{Examples}\\
                \midrule
                & \textbf{REF:}  \it{\textbf{she} is so cute}\\
                0.0112 & \textbf{A:}    \it{\textcolor{red}{heat} is so cute}\\
                0.0031 & \textbf{D:}    \it{\textcolor{blue}{he} is so cute}\\
                \midrule
                & \textbf{REF:}  \it{we \textbf{hitting that} new \textbf{club} tonight girl}  \\
                0.0219 & \textbf{A:}    \it{we \textcolor{red}{had not} new \textcolor{red}{clubs} tonight girl}  \\
                0.0167 & \textbf{D:}    \it{we \textcolor{blue}{had} new \textcolor{blue}{clubs} tonight girl}\\
                %\midrule
                %& \textbf{REF:}  \it{it's getting late what time are the \\ & \textbf{repairmen} supposed to come} \\
                %0.0056 & \textbf{A:}    \it{it's getting late what time are the \\ & \textcolor{red}{repairments} supposed to come}\\
                %0.0011 & \textbf{D:}    \it{it's getting late what time are the \\ & \textcolor{blue}{repairman} supposed to come} \\
                %\midrule
                %& \textbf{REF:}  \it{\textbf{oh} what exactly she said} \\
                %0.0064 & \textbf{A:}    \it{what exactly she said} \\
                %0.0024 & \textbf{D:}    \it{\textcolor{blue}{uh} what exactly she said}  \\
                %\midrule
                %& \textbf{REF:}  \it{it is very quiet over here \textbf{uh} the streets are empty}\\
                %0.0027 & \textbf{A:}    \it{it is very quiet over here \textcolor{red}{on} the streets are empty}  \\
                %0.0001 & \textbf{D:}    \it{it is very quiet over here \textcolor{blue}{um} the streets are empty} \\
                %\midrule
                %& \textbf{REF:}  \it{we got to go get some dinner \textbf{soon} }  \\
                %0.0075 & \textbf{A:}    \it{we got to go get some dinner \textcolor{red}{son} }   \\
                %0.0056 & \textbf{D:}    \it{we got to go get some dinner \textcolor{blue}{some} } \\
                %\midrule
                %& \textbf{REF:}  \it{\textbf{this} stay at home order is really taking} \\ & a toll on me do you have some time to video chat \\
                %0.0012 & \textbf{A:}    \it{\textcolor{red}{does} stay at home order is really taking }\\ & a toll on me do you have some time to video chat \\
                %0.0004 & \textbf{D:}    \it{\textcolor{blue}{the} stay at home order is really taking} \\ & a toll on me do you have some time to video chat  \\
                \midrule
                & \textbf{REF:}  \it{\textbf{aw} you are all so sweet } \\
                0.0057 & \textbf{A:}    \it{\textcolor{red}{or} you are all so sweet }  \\
                0.0050 & \textbf{D:}    \it{\textcolor{blue}{oh} you are all so sweet } \\
                %\midrule
                %& \textbf{REF:}  \it{\textbf{all guys} we should not do that}\\
                %& \it{because of the coronavirus is crazy right now }             \\
                %0.0018 & \textbf{A:}   \it{all \textcolor{red}{guides} we should not do that} \\
                %& \it{because of the coronavirus is crazy right now }           \\
                %0.0012 & \textbf{D:}   \it{\textcolor{blue}{oh} guys we should not do that} \\
                %& \it{because of the coronavirus is crazy right now} \\
                \bottomrule
            \end{tabular}
            }
        \end{table}
        
\section{Conclusion and Future Work}
    \label{sec:conclusion}
    In this work, we propose a novel Semantic Distance (SemDist) as an alternative evaluation metric for ASR systems, capable of measuring the semantic correctness. The SemDist measures the semantic distance between the reference and hypothesis in the embedding space by using the state-of-the-art pre-trained deep contextualized language model, RoBERTa. We demonstrate the effectiveness of our metric on various NLP downstream tasks, including named entity recognition, intent recognition, and semantic parsing. In future, we plan to explore how our Semantic Distance can be used to train ASR systems, as an additional objective.  
    
\newpage
\bibliographystyle{IEEEtran}
\bibliography{mybib}

\end{document}